\documentclass{llncs}
\usepackage[toc,page]{appendix}
\usepackage[switch]{lineno}
\usepackage{listings}
\modulolinenumbers[2]
\let\oldequation\equation
\let\oldendequation\endequation

\renewenvironment{equation}
  {\linenomathNonumbers\oldequation}
  {\oldendequation\endlinenomath}

\usepackage[english]{babel}
\usepackage{graphicx}
\usepackage{color}
\usepackage{tikz}
\usetikzlibrary{shapes.geometric, arrows.meta, positioning}
\usepackage{pgf-pie}
\usepackage[normalem]{ulem}

\newcommand{\corr}[1]{\textcolor{black}{#1}}  
\makeatletter
  \newcommand\figcaption{\def\@captype{figure}\caption}
  \newcommand\tabcaption{\def\@captype{table}\caption}
\makeatother

\definecolor{naranja}{RGB}{255,128,0}

\usepackage[T1]{fontenc}
\begin{document}

\title{The impact of LLaMA fine tuning  on hallucinations for name entity extraction in legal documents}
%
%
\author{Francisco Vargas\inst{1,2}\and
Alejandro González Coene\inst{1,2} \and
Gaston Escalante\inst{1,2}\and Exequiel Lobón\inst{1,2}\and Manuel Pulido\inst{1,3}}
%

\institute{FaCENA, Universidad Nacional del Nordeste, Corrientes
 \and
Legalhub S. A., Buenos Aires
\and
Instituto de Modelado e Innovación Tecnológica, CONICET\\
\email{franciscopvargas98@gmail.com}}
\maketitle              
\begin{abstract}
The extraction of information about traffic accidents from legal documents is crucial for quantifying insurance company costs. Extracting entities such as percentages of physical and/or psychological disability and the involved compensation amounts is a challenging process, even for experts, due to the subtle arguments and reasoning in the court decision. A two-step procedure is proposed: first, segmenting the document identifying the most relevant segments, and then extracting the entities. For text segmentation, two methodologies are compared: a classic method based on regular expressions and a second approach that divides the document into blocks of n-tokens, which are then vectorized using multilingual models for semantic searches (\textit{text-embedding-ada-002/MiniLM-L12-v2}). Subsequently, large language models (LLaMA-2 7b, 70b, LLaMA-3 8b, and GPT-4 Turbo) are applied with prompting to the selected segments for entity extraction. For the LLaMA models, finetuning is performed using LoRA. LLaMA-2 7b, even with zero temperature, shows a significant number of hallucinations in extractions which are an important contention point for named entity extraction. This work shows that these hallucinations are substantially reduced after finetuning the model. The performance of the methodology based on segment vectorization and subsequent use of LLMs significantly surpasses the classic method which achieves an accuracy of 39.5\%. Among open-source models, LLaMA-2 70B with finetuning achieves the highest accuracy 79.4\%, surpassing its base version 61.7\%. Notably, the base LLaMA-3 8B model already performs comparably to the finetuned LLaMA-2 70B model, achieving 76.6\%, highlighting the rapid progress in model development. Meanwhile, GPT-4 Turbo achieves the highest accuracy at 86.1\%.


\end{abstract}
\clearpage

%
%
\title{El impacto del ajuste fino de LLaMA en las alucinaciones para la extracción de entidades nombradas en documentos legales}
%
%
\author{Francisco Vargas\inst{1,2}\and
Alejandro González Coene\inst{1,2} \and
Gaston Escalante\inst{1,2}\and Exequiel Lobón\inst{1,2}\and Manuel Pulido\inst{1,3}}
%

\institute{FaCENA, Universidad Nacional del Nordeste, Corrientes
 \and
Legalhub S. A., Buenos Aires
\and
Instituto de Modelado e Innovación Tecnológica, CONICET\\
\email{franciscopvargas98@gmail.com}}
\maketitle              

\begin{abstract}
La extracción de información sobre accidentes de tráfico a partir de documentos legales es crucial para cuantificar los costes de las aseguradoras. Extraer entidades como los porcentajes de discapacidad física o psicológica y las indemnizaciones implicadas es un proceso complejo, incluso para expertos, debido a la sutileza de los argumentos y razonamientos de la sentencia judicial. Se propone un procedimiento en dos pasos: primero, segmentar el documento identificando los segmentos más relevantes y, posteriormente, extraer las entidades. Para la segmentación de texto, se comparan dos metodologías: un método clásico basado en expresiones regulares y un segundo enfoque que divide el documento en bloques de n-tokens, que posteriormente se vectorizan mediante modelos multilingües para búsquedas semánticas (text-embedding-ada-002/MiniLM-L12-v2). Posteriormente, se aplican modelos de lenguaje grandes (LLaMA-2 7b, 70b, LLaMA-3 8b y GPT-4 Turbo) con prompts a los segmentos seleccionados para la extracción de entidades. Para los modelos LLaMA, se realiza un ajuste fino mediante LoRA. LLaMA-2 7b, incluso a temperatura cero, presenta un número significativo de alucinaciones en las extracciones, lo cual constituye un importante punto de contención para la extracción de entidades nombradas. Este trabajo demuestra que estas alucinaciones se reducen sustancialmente tras el ajuste fino del modelo. El rendimiento de la metodología basada en la vectorización de segmentos y el posterior uso de LLM supera significativamente al método clásico, que alcanza una precisión del 39,5 \%. Entre los modelos de código abierto, LLaMA-2 70B con ajuste fino alcanza la mayor precisión, con un 79,4 \%, superando a su versión base con 61,7 \%. Cabe destacar que el modelo base LLaMA-3 8B ya presenta un rendimiento comparable al del modelo LLaMA-2 70B ajustado, alcanzando un 76,6 \%, lo que demuestra el rápido progreso en el desarrollo del modelo. Por otro lado, GPT-4 Turbo alcanza la mayor precisión, con un 86,1 \%.


\end{abstract}

\section{Introduction}
The judicial proceedings in Argentina are composed of several communication steps between the plaintiff, defendants, and judicial entities that end in a court decision. This work deals with documents of civil jurisdiction related to lawsuits for damages. 
In the civil jurisdiction, insurance companies and law firms require the processing of a large number of court decisions to quantify costs and make decisions on the procedure to follow depending on the case.

To quantify the damage for each injury over the plaintiff, a medical expert has a set of rules to assign the percentage of disability for each injury. This disability caused in the actor can be of two types, physical or psychological. The percentage value assigned to the degree of injury of an actor depends on the specific case and is not fixed. Moreover, each assigned value has an indemnity amount to be paid to the actor. The data analysis can contribute to identifying trends in the relations between the indemnity amount and the percentage of disability. Both the defendant and plaintiff can leverage this trend to ask for a better indemnity or appeal with a higher probability of being approved. To carry out the trend estimation it is necessary to perform an automatic recognition to process a large amount of decision courts. Furthermore, differences between judges or courthouses in the treatment of the cases may be an important piece of information for decision-making. This problem has a big challenge because the entities to be recognized are sparse and the text is long. Several times similar entities can be found in different contexts and entities can be related to references to other cases that are not necessarily related to the current case and this can interfere with the quality of the extraction process.

Text segmentation is required to reduce interferences in the named entity extraction to choose the more relevant sections or contexts. This could be conducted by leveraging a distinctive feature of legal documents which is their structure. This allows us to make a hierarchical segmentation without the need for a machine learning model\cite{BAYOMI18.806}. An essential condition to choose the relevant segments with the entities to extract and the context is the existence of keywords and symbols. This rule based on keywords, is simple but not definitive when selecting the segments. Indeed, important segments may be ignored because of the absence of the fixed keywords used in the method. Therefore, to improve the selection of potential text segments that contain the type of disability and compensation amount, we propose and evaluate a vectorial similarity search based on \textit{text-ada-002} over the potential segments conforming the entire legal document and a query-based search \cite{alemi2015textsegmentationbasedsemantic}. The two methods of segmentation based on keywords and model-based segmentation are assessed in this work. An alternative methodology could be to use segmentation based on topics. This involves the training of a model to conduct two depending tasks, the segmentation and the labelling of each segment with the corresponding topic \cite{barrow-etal-2020-joint}. This may be of interest if there are several types of entities to extract which can be found in different sections of the document and there are differences in the order of the sections of the documents. For the current work, we focus on a few entities of interest, and topic segmentation was not needed. Another relevant aspect of the segmentation is the length. If the candidate segments are too short, these are not going to have the required context to identify if the found named entities are the final decisions or are similar entities part of the arguments and reasoning that lead to the final decision. In this sense, the typical writing of legal documents is characterized by small variance and highly structured so that a long context is required to identify the correct named entities.

Named entity recognition (NER) is a usual task in data science and is widely applied in the legal field \cite{Vardhan}. The entity recognition usually assumes that the frequency of entities (e.g. the name of a person, judge, or actor) that one is interested in extracting is low in a document and even more between documents \cite{Karaa2011NamedER}. In our case, because we are required to recognize percentual amounts and injure compensation amounts, these can be present repeatedly in a document and they are present in all documents. However, these entities represent different magnitudes in each context and we require to distinguish them using the context in which the entity is found.

The latest advancements in natural language processing have revolutionized text understanding through the use of large language models (LLMs), such as GPT and LLaMA \cite{ROUMELIOTIS2024100056}. These models allow us to optimize their behavior through \textit{prompt engineering} without requiring specific training for a given task.
In production environments, the cost-benefit ratio between a fine-tuned model versus a not fine-tuned one must be considered. Additionally, it is important to weigh the advantages of using open models like LLaMA-2 and LLaMA-3 compared to commercial and closed models like GPT-3, GPT-3.5 and GPT-4, which are updated over time and whose code and training data remains unknown \cite{chen2023chatgpts}.

In this work, we propose an approach to address the identification of named entities in legal documents. A two-stage procedure is introduced. In the first stage, the most relevant text segments are identified. To achieve this, different text segmentation methods are evaluated, including those based on regular expressions and vectorization (Section \ref{metodo:segmentacion}). In the second stage, large language models, such as GPT-3.5, LLaMA-2, and GPT-4, are used for entity extraction, and their performance is compared with keyword-based methods using regular expressions (Section \ref{metodo:LLM}). The results of applying these methodologies and comparing their performance are presented in Section \ref{resultados}. The application of prior knowledge on the texts to improve the extraction process is also evaluated.

\section{Methodologies}
\subsection{Database}\label{preprosylimpieza}
The documents used in this work are first-instance judicial rulings related to civil damage lawsuits. These documents are available on the platform of the Argentine National Judiciary (PJN)\cite{poderjudicial} in digitized format. They are public and freely accessible.

The documents were collected using a scraper from the PJN website. Within the documents, rulings related to accidents were selected. The generated database contains a set of 650 first-instance judicial rulings, whose final judgments were issued in October 2023.

Before the analysis, a preprocessing and data cleaning phase is conducted consisting of several steps. First, the documents are classified according to their format, separating PDFs with extractable text from the documents that consist of images or scanned content, discarding those in the latter category. Then, the PyPDF2 tool \cite{PyPDF2} is used to extract the text from the selected documents, and regular expressions are applied to remove irrelevant or undesired information. Among these, the codes of the rulings that are repeated in the header of each page are removed.

As a final step, a second document-level filtering is performed to isolate rulings that fall within the defined scope, specifically traffic accidents with injured individuals but without fatalities. 
This filtering involved analysing the headers of each ruling and, using domain knowledge, defining keywords that must or must not appear for each ruling to fall within the defined field of study. Using these regular expressions, each header of the rulings is classified the ones within the scope of this study are selected.

Applying the previous filters to the database of 650 rulings resulted in 278 that fall within the scope of this work. For the generation of the labelled database which is used for finetuning a hybrid procedure was used, wherein a language model was initially employed for label extraction, followed by human intervention for quality control. This process is described in Section \ref{generacion_etiquetas}.

\subsection{Document Segmentation}\label{metodo:segmentacion}

\subsubsection{Segmentation Method with RegEx}
Segmentation by means of regular expressions (RegEx) can be used to divide judicial rulings into smaller sections to fit the maximum context processed by language models. The goal is to identify text segments within the judicial rulings that contain information about physical or psychological disabilities, as well as the percentage of disability and associated amounts. The main entities of interest are summarized in Table \ref{tab:entidades}\label{entidades}.

\begin{table}
    \centering
    \begin{tabular}{|c|c|}
    \hline
    Entity & Entity Properties\\
    \hline
    Physical Disability & Percentage of disability | Compensation amount\\
    Psychological Disability & Percentage of disability | Compensation amount\\
    Moral Damage Amount & Compensation amount\\
    \hline
    \end{tabular}
    \caption{Entities for extraction.}
    \label{tab:entidades}
\end{table}

The chosen RegEx segmentation process consists of:
\begin{itemize}
    \item \textbf{Search for the '\%' symbol within the text:} The purpose of this search is to locate potential indicators of disability percentage in the text. 
    Regular expression: \texttt{r"[\textbackslash w\textbackslash d\textbackslash s\textbackslash n,.]\{0,1\}\%"}
    \item \textbf{Context determination:} For each percentage match, a context is established around it. This context is defined by a window of 500 characters before and after the position of the match, resulting in a context of approximately 320 tokens.
\end{itemize}

\subsubsection{Vectorization and Semantic Search Segmentation Method:}\label{subsubsection:segmentacion}
Document segmentation using vector search is a process in which the content of a text is represented within a vector space using a transformer-based model \cite{MiniLM}. Through this "vectorizer", each segment of the document is represented by a limited set of vectors. Within this vector space, searches can be conducted by similarity measures, for instance cosine similarity. 
In this work, the \textit{paraphrase-multilingual-MiniLM-L12-v2} embeddings model from Huggingface \cite{MiniLM} was employed.

This model is particularly relevant due to its ability to handle multiple languages, including Spanish, 
 by its high efficiency, and its capability of vectorizing text at the token level, making it an ideal choice for tasks requiring fine-grained semantic representations across diverse contexts.

The segmentation process is carried out through the following steps:
\begin{itemize}
    \item \textbf{Text division into blocks:} Text Division into Blocks: The plain text of the document is divided into fixed-size segments of 120 tokens. This size was chosen due to the
    limitations of the embeddings model \cite{MiniLM} (maximum of 124 tokens) used and to facilitate feature extraction without compromising semantic accuracy.
    \item \textbf{Text block vectorization:}  
    These text blocks are vectorized using the embeddings model and stored with the FAISS library \cite{douze2024faiss} (\textit{Facebook AI Similarity Search}), which generates specific indices to optimize the search process.
    \item \textbf{Vector search:}  
    A vector search is performed on the vectors generated from a query, which is compared to the vectorized text blocks.
    \item \textbf{Query generation for the search:}  
    To define the query text used in cosine similarity search, a Tf-idf (\textit{Term Frequency – Inverse Document Frequency}) is applied to the expected text blocks. This process identifies key terms to use in the search, improving accuracy by aligning the query closely with the intended content.
    \item \textbf{Contextual expansion of text blocks:}  
    From the blocks retrieved through semantic search, context expansion is carried out by concatenating the retrieved block with its preceding and succeeding blocks. This results in new vectors that combine all three blocks (the retrieved block, plus the preceding and succeeding blocks), creating a 360-token text block that enhances contextual understanding for extraction.
\end{itemize}

\subsection{Entity Extraction with Regular Expressions}

\label{regex_NER} \corr{The classical method used for entity extraction is the use of regular expressions. This approach is implemented in current systems, making it the baseline reference to measure the impact of using language models.} For this extraction method to work, a single relevant segment must be predefined, within which the following is performed:

\begin{itemize}

    \item \textbf{Keyword search to identify the type of disability:} Within the context, regular expressions are used to search for keywords that indicate the type of disability associated with the identified percentage. This includes words such as 'physical' or 'psychological.' Depending on the keywords found, the type of disability associated with the percentage is determined.

   \item \textbf{Disability percentage extraction:} In addition to identifying the type of disability, regular expressions are used to extract the percentage of disability from the context. This is achieved by searching for numbers followed by a percentage sign within the context.

   The regular expression is: r"(\textbackslash d+(?:,\textbackslash d+)?(?:.\textbackslash d+)?)\textbackslash s*\%"

    \item \textbf{Amount extraction:} Similarly, amounts are extracted by identifying the dollar sign and numbers within the relevant segment.

\end{itemize}

    
\subsection{Using langauge models for entity extraction}\label{metodo:LLM}
In the last years, several language models have been developed with increasing performance. For this work, we examined different versions of the open source LLaMA model \cite{touvron2023llama} developed by Meta. The choice was made because it is one of the most powerful open source language models and many works have tested it for different applications including legal documents. We have also used in this work the models GPT-3.5 Turbo and GPT-4 Turbo as baselines to compare the performance of Llama in Spanish legal documents.




\subsubsection{General Methodology}
\label{RAG_metodologia}
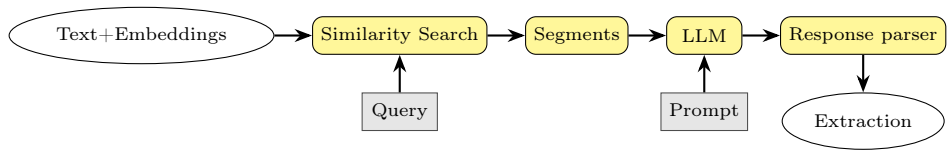
\begin{figure}
\centering
\begin{tikzpicture}[
    font=\scriptsize,
    node distance=0.25cm and 0.5cm, 
    box/.style={rectangle, draw=black, fill=yellow!50, rounded corners, minimum height=0.5cm, minimum width=1cm}, 
    input/.style={ellipse, draw=black, minimum height=0.75cm, minimum width=0.60cm},
    aux/.style={rectangle, draw=black!70, fill=gray!20, minimum height=0.5cm, minimum width=1cm},
    arrow/.style={thick,-{Stealth}}
]

\node (input) [input] {Text+Embeddings};
\node (search) [box, right=of input] {Similarity Search};
\node (query) [aux, below=0.5cm of search] {Query};
\node (segments) [box, right=of search] {Segments};
\node (llm) [box, right=of segments] {LLM};
\node (prompt) [aux, below=0.5cm of llm] {Prompt};
\node (parser) [box, right=of llm] {Response parser};
\node (output) [input, below=0.5cm of parser] {Extraction};

\draw [arrow] (input) -- (search);
\draw [arrow] (search) -- (segments);
\draw [arrow] (segments) -- (llm);
\draw [arrow] (llm) -- (parser);
\draw [arrow] (parser) -- (output);
\draw [arrow] (query) -- (search);
\draw [arrow] (prompt) -- (llm);
\end{tikzpicture}
\label{flujo_extraccion}
\caption{Complete extraction pipeline.}
\end{figure}

The general process for entity extraction with a language model uses a {\em Retrieval-Augmented Generation} (RAG) methodology as summarized in Figure \ref{flujo_extraccion}. Given the document, a set of optimized \textit{queries}, and a \textit{prompt}, the segmenter is first used with the \textit{queries} to divide the document's text into different blocks, generating their \textit{embeddings} and storing them. In a second stage, the language model is fed with each block, their corresponding vectors, and an optimized \textit{prompt} to extract the entities and their properties.



\subsection{Training dataset}\label{generacion_etiquetas}

 There are 278 filtered and preprocessed documents available (see Section \ref{preprosylimpieza}), which are useful for training the model. To perform the training of the model, a database is needed that contains the relevant segments of each document labelled with the entities to be extracted. For the database labelling process, it was proposed to use an LLM with subsequent human intervention.

First, we focus on obtaining an optimal prompt for entity extraction. We refer to the prompt as the natural language instruction given to the model. For this optimization, there are different prompting techniques (for example, think step by step \cite{kojima2023large}, follow specific formats to facilitate understanding and the structure of the text, among others), which are used to create efficient instructions and achieve the best responses from large language models. The prompts used can be found in \footnote{http://github.com/Fran-98/JAIIO-ASAID-141}.

The model used for this initial labelling of the database was OpenAI's GPT-4 Turbo. Labelling consists of using an RAG, in which the document segments are extracted, as described in Section \ref{subsubsection:segmentacion}, using it as input to the LLM along with the prompt so that the LLM can extract the entities from the relevant information provided. In a second stage, manual quality control is performed on the resulting labels obtained with the LLM to verify that they are indeed the entities of the document. This mixed labelling allows us to save considerable man-hours when correcting the data, as only poorly performed extractions by the model need to be corrected.

The labelled database is then used for two purposes; part of it is used for fine-tuning. The other part is reserved to have a reference (test set) to evaluate the performance of the different LLMs in the various experiments we conduct.

The segmenter used for labelling the database corresponds to OpenAI's embeddings model \textit{text-embedding-ada-002}. With these segments, the language model is fed to extract the entities. Once all documents were processed, as mentioned, a manual data cleanup was performed on the total number of files. This involved reviewing each of the PDFs and the extracted entities, which are the physical, psychological, and psychophysical disabilities along with their characteristics (compensation amount and percentage) and the compensation amount for moral damages, correcting the labels as necessary.


From this manual process, the correct labels that the model must provide for each extraction are obtained. In this way, the labelling of the complete dataset is achieved, resulting in a total of 1120 samples, corresponding to the 4 entities in the 278 documents. This sample set is referred to as \textit{Dataset 1}. We assume the hypothesis that the sample of one entity will contribute to the trained model's ability to recognize other similar or the same type in new texts.

The samples from the manually corrected dataset are still not perfect for training. There are labels that would be impossible to extract as they are not found in the segments provided by the segmenter. In these cases, the label should be empty. Therefore, a second dataset was generated. Manual searches for the entities were conducted within the relevant segments. If the entities were not found in the segment, those samples were discarded. An option could have been choosing to use those samples and label them as empty; for the construction of this dataset, it was decided to discard them. After this clean-up, 861 samples remained, and this set is referred to as \textit{Dataset 2}.

As a test database, 30 documents within the same domain were used, resulting in 120 samples. These documents do not belong to the training set, and their preparation was similar to the manual clean-up performed for the training \textit{Dataset 2}.

\subsection{Finetuning}

Due to the VRAM restriction of 96Gb available (corresponding to 2xNvidia A40 GPUs), it was required to use quantization techniques for the training. This is the process by which the precision of the model weights is reduced so that they occupy less memory space and require fewer resources for training and inference. Quantization also increases inference speed. In this work, the \textit{bitsandbytes} library was used to perform the quantization from 32-bit floating point to 8-bit integer, i.e., FP32 $\rightarrow$ int8. The degradation of the weights does not significantly affect the accuracy of the inference and allows the use of larger parameter models, for example, 70b, thus achieving better results than if a 13b model is used without any type of quantification \cite{jin2024comprehensive}.

Furthermore, the \textit{finetuning Quantized Low Rank Adaptation} method \cite{dettmers2023qlora}, QLoRA, was used which is a method derived from \cite{hu2021lora} with added quantization. LoRA introduces the possibility of training adapters that contains far fewer parameters than the base models and can be integrated into them to achieve effective finetuning using many fewer resources than if we attempted to update all model weights. These methods belong to the family of PEFT methods (\textit{Parameter-Efficient Fine-Tuning}) \cite{xu2023parameterefficient}. QLoRA in this case allowed us to perform the finetuning with the available computational resources.

Two trainings were conducted, one for each dataset, leaving 10\% of each dataset as a validation set. The trained layers were the attention layers, with a learning rate of 5e-5, for 3 epochs and a LoRA range of 8. The most suitable checkpoint was selected for each dataset, aiming for the optimal point where the model has generalized the training information but does not exhibit over-fitting.

\section{Results}\label{resultados}
\subsection{Segmentation}
During the document segmentation phase, it is essential to assess the effectiveness of semantic search and its ability to correctly identify the necessary segments for data extraction by the language model. Without this information, the model would not be able to extract the required entities, as it would lack access to the necessary data. A quality assessment (QA) of the segmentation was conducted on the sentences. This evaluation allowed for determining whether the semantic search effectively retrieves the segments containing the correct data for subsequent extraction. 
The QA results, performed on a test set of 30 sentences, showed an accuracy rate of $83.33\%$ using the "\textit{sentence-transformer/paraphrase-multilingual-MiniLM-L12-v2}" model \cite{MiniLM} and an accuracy rate of $80.91\%$ with the \textit{text-embedding-ada-002} embeddings model from OpenAI.

\begin{figure}
\centering
\includegraphics[width=1\textwidth]{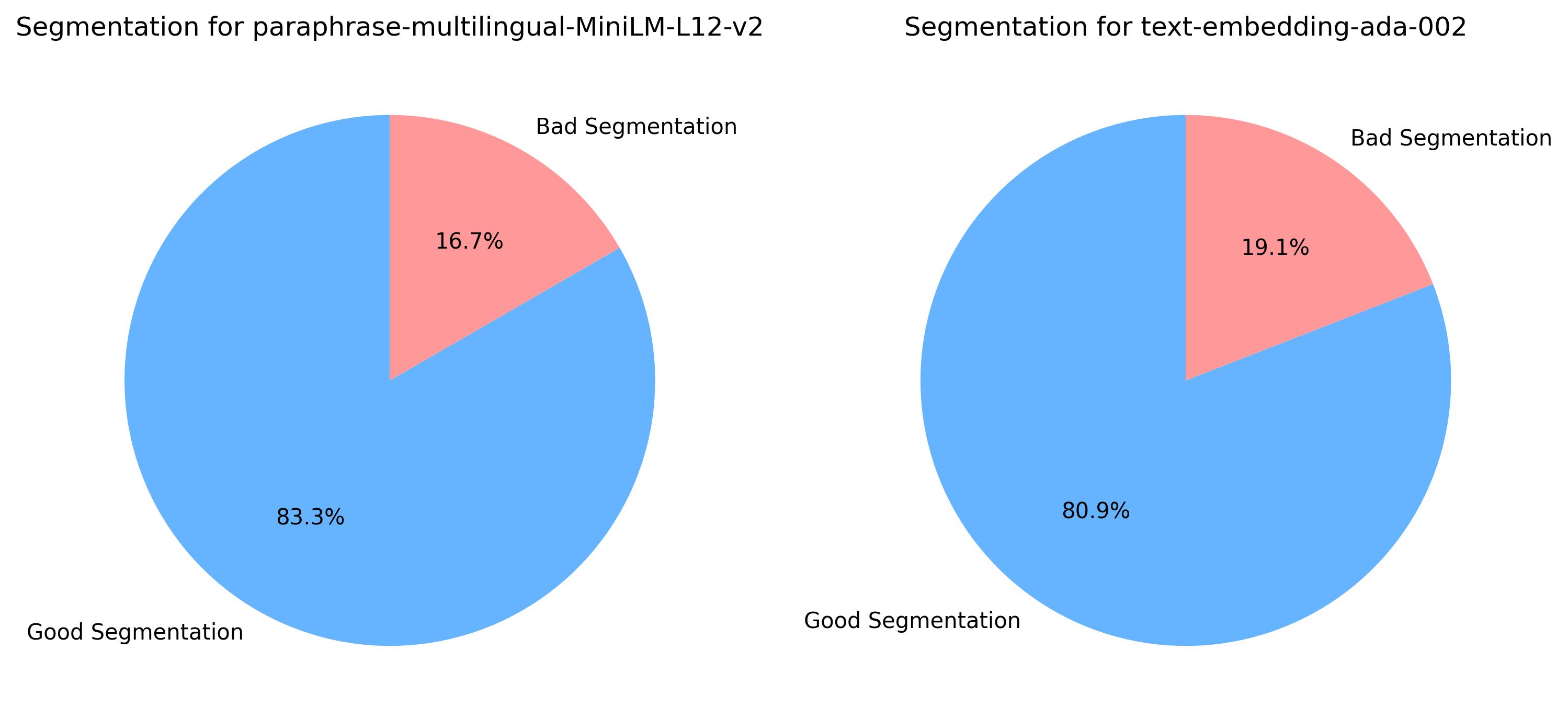}
\caption{Segmentation results by model.} \label{resultados_segmetantacion}
\end{figure}

 \subsection{Extraction results}

\begin{figure}
\centering
\includegraphics[width=1\textwidth]{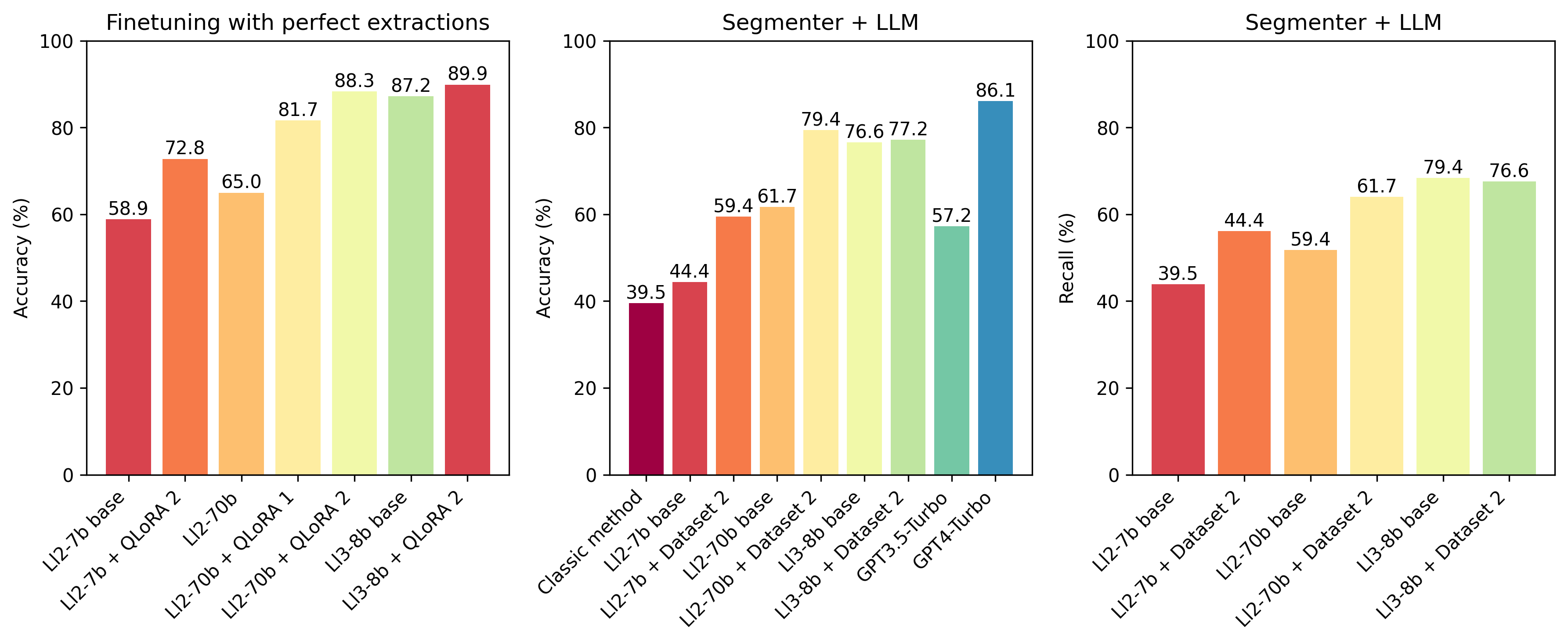}
\caption{a) Finetuning impact on accuracy for each model considering a perfect segmentation. b) 
 Accuracy of the complete model including the segmentation and the entity extraction language model. c) Recall from the entire pipeline taking into account segmenter and model.}\label{resultados_finetuning_modelos}
\end{figure}

The results obtained using the segments from the \textit{text-embedding-ada-002} model with the finetuned models of LLaMA-2 7b, LLaMA-2 70b, and LLaMA-3 8b, are shown in Graph \ref{resultados_finetuning_modelos} a).

Significant improvements in performance were observed when tuning with the specialized datasets. On the other hand, there is a gain of 9.4\% in accuracy from \textit{Dataset 1} to \textit{Dataset 2}. The latter significantly outperforms the former; it manages such an accomplishment despite having fewer samples. The difference in performance was made possible by the cleaning performed. This highlights the importance of having a quality dataset when training any model.

The best result is achieved by the finetuned version of LLaMA-3 8b with \textit{Dataset 2}, but it shows a small improvement of 2.7\% over its base model. As a result, it is evident that the finetuning process was less impactful on the third iteration of the llama family models, contrary to what was achieved with its predecessors.

The LLaMA-2 70b model tuned with \textit{Dataset 2} shows a 17.7\% improvement over the base model. The finetuning on LLaMA-2 7b has a lesser impact with a difference of 13.9\%.

Comparing only the base versions of each model the evolution of the LLaMA family is evident. The best performing base model is LLaMA-3 8b which is on par to the best results obtained, outperforming LLaMA-2 70b by 22.9\% and the 7b version by 29.0\%.

The accuracies shown in Figure \ref{resultados_finetuning_modelos} a) correspond to a perfect segmenter; however, when performing inference automatically, it is likely that the segmenter selects some segments incorrectly. The effectiveness of the entire process, including the segmenter and the model, is presented in Figure \ref{resultados_finetuning_modelos} panel b).The recall calculated for the tuned models is observed in figure \ref{resultados_finetuning_modelos} panel c), which is lower than the measured accuracy, indicating that the model is more conservative when delivering a value and attempts to maintain high reliability when making an extraction. This allows us to be confident in its results, but at the same time, a substantial part of the information that could be used is lost.

The results obtained with the proposed method far exceed those obtained with the classic method, RegEx. This method yields a performance of 39.5\% on the complete pipeline. Meanwhile, LLaMA-2 70b with finetuning achieves a performance of 79.4\% and LLaMA-3 8b 77.2\%, slightly lower than the latter. The performance obtained by the LLaMA family is significantly higher than GPT3.5-Turbo but lower than GPT4-Turbo, which achieves 86.1\%.

On the other hand, there is clear evidence of a strong decrease in performance of 19.4 \% in the case of the base LLaMA-2 7b model when the segments are not perfect compared to when the segments are perfect (Figure \ref{resultados_finetuning_modelos}). In these situations, when the entity is not correctly identified in the segment, the model tends to generate invalid responses. This is because when the model does not find the correct entity in the segment, it tends to hallucinate a non-existent response in the segment. This result motivated the need to conduct an evaluation of hallucinations in the models.

\subsection{Hallucinations}

Hallucinations are a major issue when using LLMs, as due to their nature, these models generate responses regardless of whether they are correct or not. The case of entity extraction is not foreign to this problem and it can occur that the extractions are affected by this eagerness to find an answer. In particular, as mentioned, the analysis of case types of the inferences of the LLaMA-2 7b model shows that hallucinations have a strong impact on its poor performance.

\begin{figure}
\centering
\includegraphics[width=0.5\textwidth]{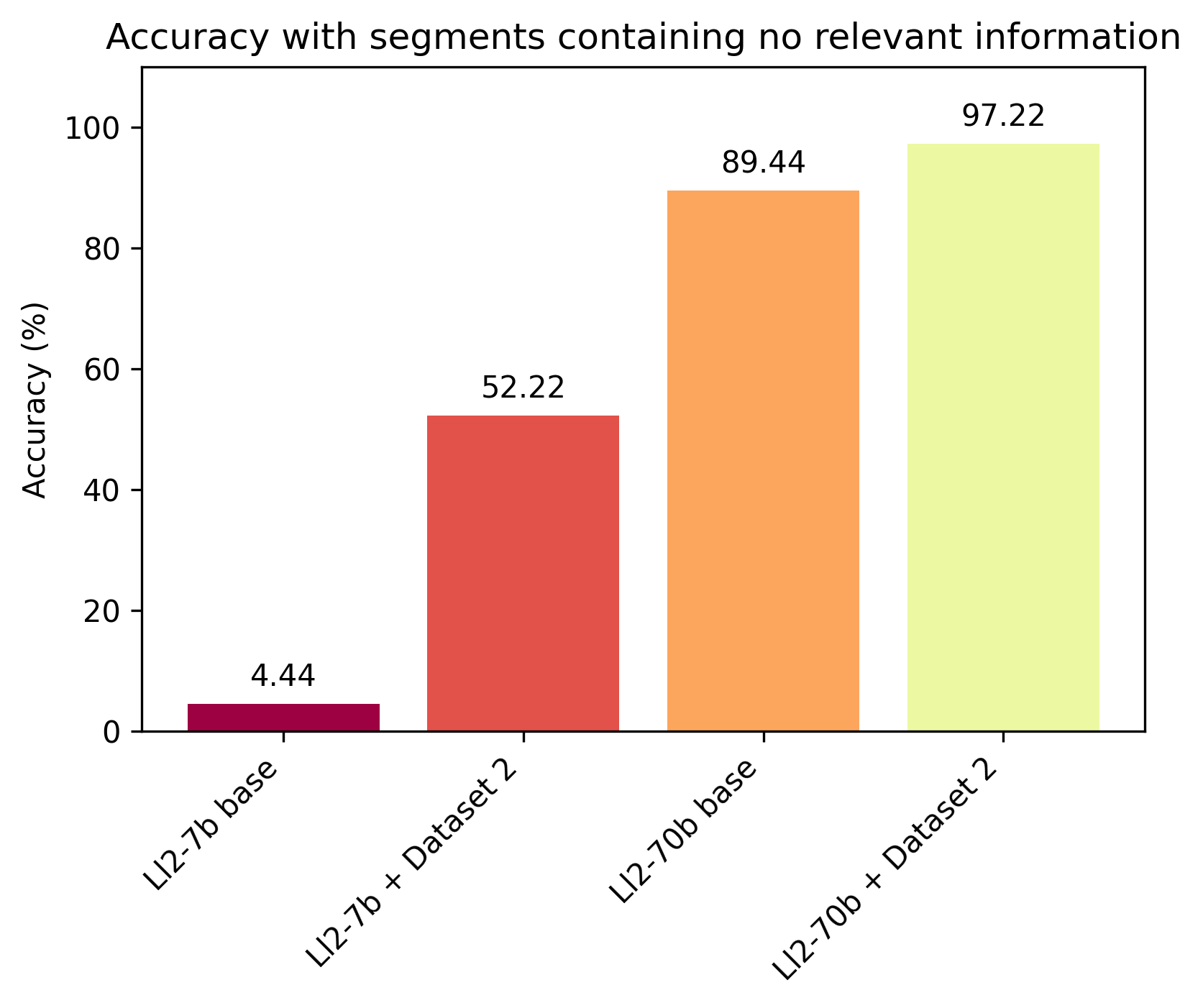}
\caption{Performance of the models with segments that do not contain the entities to extract.} \label{resultados_alucinaciones}
\end{figure}

To quantify this phenomenon, a dataset was created with the discarded segments to form  \textit{Dataset 2}, where the entities to be extracted are not present. An amount of segments equivalent to those obtained from 30 judicial sentences was used. The results of this experiment are shown in Figure \ref{resultados_alucinaciones}. The LLaMA2-7b model performs very poorly when the information is not present within the segment. A notable result is the significant improvement with the finetuning of the LLaMA2-7b model, showing a difference of 47.78 \% compared to its base version. This difference is much smaller when compared to the case of the LLaMA2 70b model. In this case, the base model has a low amount of hallucinations as it has a better ability to follow instructions and understand texts, with a difference of 7.78 \%. In both cases, the improvement from finetuning is remarkable in reducing hallucinations.

To reduce hallucinations and their impact, a hallucination detection method was evaluated using the probabilities associated with the tokens generated by the model. The mitigation method is based on the idea that the probability of a token is a quantification of the certainty the model has about generating the selected token as the next one. This probability or certainty is conditioned by the prompt and the context of the text that has already been generated.

If we consider the probabilities as $p_1, p_2, ..., p_n$, according to \cite{varshney2023stitch} the most suitable method for detecting a hallucination is by taking the minimum probability within the set, $p_{min} = \min(p_1, p_2, ..., p_n)$. We can define a minimum probability threshold, $ p_{min}< p_u$  such that any generation containing a probability in its set that is lower than this threshold will be considered a potential hallucination.

If we take the probabilities as $p_1, p_2, ..., p_n$, according to \cite{varshney2023stitch}, the most suitable method for detecting a hallucination is by taking the minimum probability within the set, $p_{min} =   ext{min}(p_1, p_2, ..., p_n)$. We can define a minimum probability threshold, $ p_{min}< p_u$, in such a way that any generation that has a probability lower than this in its probability set will be considered a potential hallucination.

The minimum threshold method for hallucination detection was analysed using the test dataset, considering both successful and unsuccessful extractions. Using the minimum threshold as a detection method, it was not possible to recognize any pattern or value for the threshold; there are cases where the model has a very low percentage but the extraction is correct, as well as the opposite occurring. Thus, we can clearly see that an LLM can be very confident about an incorrect answer. In conclusion, it is determined that for the case of entity extraction, this method alone is insufficient to detect incorrect extractions.

\subsection{Statistics of the extractions}
\label{resultados:utilidad}

Once the extractions have been made, we proceed to perform the statistics of the extracted entities. This information is useful for insurers as well as for the plaintiff and the defendant, who can use this information to request or appeal for compensations. One of the amounts estimated from the extracted entities is the point value, $PV$. This is a calculation used to estimate the amount assigned by judges for each percentage point of incapacity of an individual. These statistics are crucial for insurers to estimate the liquidity needed to cover lawsuits as well as to request or appeal for compensations. To determine the $PV$, we consider the amounts attributed to psychological incapacity $PSI_a$ and its percentage $PSI_p$, the amounts of physical incapacity $PI_a$ and its percentage $PI_p$, and the amount attributed to moral damage $MD_a$, resulting.
\begin{equation}\label{valor_punto}
    PV = \frac{PSI_a}{PSI_p} + \frac{PI_a+DM_a}{PI_p},  
\end{equation}

where in the last term for the amount of moral damage it is assumed that the percentage of physical incapacity is more determining and significant than the percentage of psychological incapacity.

\begin{figure}
\centering
\includegraphics[width=1\textwidth]{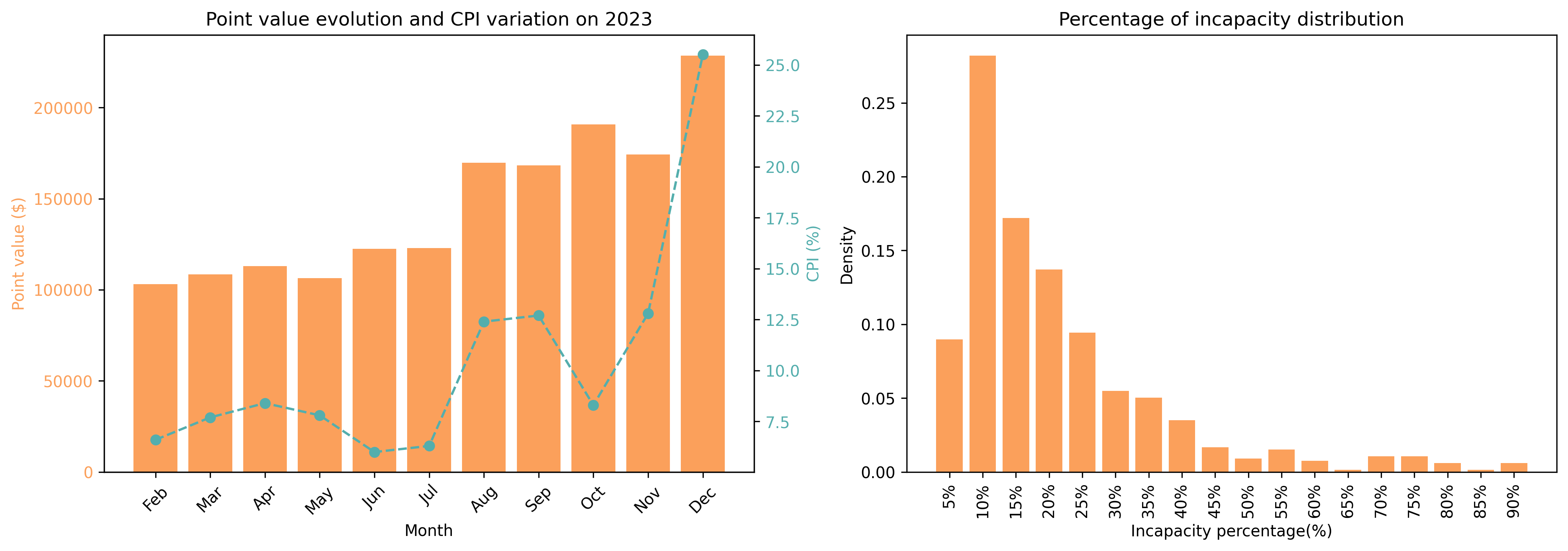}
\begin{minipage}[c]{0.47\linewidth}
        \caption{Variation of the point e value
        CPI during 2023, obtained through the procedure described in Figure \ref{flujo_extraccion}}
        \label{grafico_valor_punto}
\end{minipage}%
\hspace{0.1in}
\begin{minipage}[c]{0.47\linewidth}
    \caption{Distribution of disability percentages by court ruling obtained from entity extraction.}
    \label{distribucion_valor_punto}
        
\end{minipage}

\end{figure}

The inference was carried out on all rulings from the year 2023 corresponding to the jurisdictions circumscribed to "Capital Federal, Buenos Aires, Argentina". From these rulings, the entities were automatically extracted following the RAG flow described in Figure \ref{flujo_extraccion} with the best-performing model, GPT-4 Turbo, and from these, the point value was estimated using (\ref{valor_punto}). In general terms, it can be observed in Figure \ref{grafico_valor_punto} based on the month of the ruling that the point value has a significant correlation with the accumulated value following the consumer price index (CPI); however, there are also particular behaviours that cannot be described through the CPI.

In this way, the calculation of the point value can be obtained for an entire year with an automatic procedure that requires an average of half an hour of computing time. The traditional procedure for obtaining the point value consisted of manually analysing each ruling to extract each entity and then performing the calculation manually, which took more than a week of work per month for an experienced lawyer in the field.

Another piece of relevant information for insurance companies and legal defence is the distribution of disability percentages. This analysis is fundamental as it allows for an understanding of how disability grades are allocated in judicial rulings, thereby providing a solid foundation for strategic decision-making. That is, by considering all rulings within a specific time period as the sample and extracting the disability percentage from each decision, a histogram can be constructed. This histogram not only illustrates the frequency of different disability levels but also reveals patterns and trends that can be used to predict future behaviours.

Considering the distribution of disability percentages, the point value, and the number of cases, these elements enable us to perform statistical estimations with objective and up-to-date information regarding the costs involved. This information is crucial for optimizing legal and financial strategies, as it provides a quantitative basis for decision-making. For instance, by analysing historical data, insurance companies can identify recurring patterns, such as the frequency with which certain disability percentages are granted in specific jurisdictions. This allows them to adjust their strategies based on economic variables and the behaviour of each jurisdiction, as well as to calculate the liquidity required to cover legal proceedings.

In this way, insurance companies can determine when it is advisable to reach a settlement before a judge issues a decision. This decision is based on a thorough analysis of the potential costs associated with an unfavourable ruling compared to the costs of an out-of-court settlement. Additionally, they can assess the advisability of awaiting a judicial decision based on previous patterns derived from the point value. The point value is a key indicator that reflects the cost of $1\%$ disability in a specific jurisdiction, based on rulings from the past month. This indicator is dynamic and is constantly updated, making it an invaluable tool for financial and legal planning.

The distribution of disability percentages in rulings obtained for the 2023 decisions is shown in Figure \ref{distribucion_valor_punto}. It is observed that $90\%$ of the rulings involve a disability of less than $30\%$, indicating that the majority of judicially resolved cases involve relatively low disability levels. On the other hand, only $7.16\%$ of the cases exceed $50\%$ disability, suggesting that more severe disabilities are less frequent but can have a significant financial impact. This type of analysis enables insurance companies to prepare for both common and exceptional scenarios, ensuring that they have the necessary resources to address any eventuality.

In summary, the study of the distribution of disability percentages, along with the point value and the number of cases, provides a solid foundation for informed decision-making. It not only enables insurance companies to optimize their legal and financial strategies but also helps them manage risks and ensure long-term financial sustainability. This data-driven approach, based on historical records and statistical analysis, is essential in an increasingly complex and competitive legal and economic environment.

\section{Conclusions}

The language model-based method proposed in this work demonstrates a significant improvement over classical methods based on regular expressions for entity extraction in judicial sentences. At the same time, we show that an open-source LLM can significantly enhance this task when trained with a relatively small database, achieving a maximum accuracy of 79.4\%, surpassing well-established commercial models like GPT-3.5 Turbo and approaching GPT-4 Turbo's performance, which stands at 86.1\%. The impact of finetuning on LLaMA-2 70b was 29.4\%, going from 58.90\% on the base model to 79.4\% obtained with the trained model. Notably, LLaMA-3 8B base performs comparably to the fine-tuned version of its predecessor (LLaMA-2), with only a 2.2\% difference. This result highlights the rapid advancements in the field, as newer models become smaller and more capable.
Finetuning significantly reduced the number of hallucinations found in the base LLaMA-2 7b model, which had the lowest performance among those tested.

The current systems in use for the identification of disabilities are based on the use of regular expressions, so this work focused on measuring the impact of using language models compared to the regular expression method. A substantial improvement was obtained with the use of large language models (GPT-4 Turbo, LlaMA-2, LLaMA-3). The analysis did not include the performance difference between large language models (LLMs) and more classical language models like BERT.

The proposed technique offers the possibility of replacing the manual work of data extraction from judicial sentences related to accidents with an automated workflow, segmenter/LLM, allowing for the analysis of a large volume of documents, thereby obtaining information that was previously unfeasible, including the distribution of probabilities of disabilities and the monthly point value, as demonstrated in Section \ref{resultados:utilidad}.

The procedure described in this work allows for the discrimination of the estimated point value by jurisdictions and even at the level of individual judges, so the general statistical analysis performed could be extended to the detection of jurisdictional anomalies or at the judge level regarding point values and/or disability percentages. On the other hand, it could potentially detect changes in the tail of the disability distribution by jurisdiction or by judge.

%
%
%
%
\vskip 4mm

{\em Acknowledgments} The training and inferencing with large language models were carried out on the computing servers of CECONEA (UNNE).


\begin{thebibliography}{10}
\providecommand{\url}[1]{\texttt{#1}}
\providecommand{\urlprefix}{URL }
\providecommand{\doi}[1]{https://doi.org/#1}

\bibitem{alemi2015textsegmentationbasedsemantic}
Alemi, A.A., Ginsparg, P.: Text segmentation based on semantic word embeddings (2015), \url{https://arxiv.org/abs/1503.05543}

\bibitem{barrow-etal-2020-joint}
Barrow, J., Jain, R., Morariu, V., Manjunatha, V., Oard, D., Resnik, P.: A joint model for document segmentation and segment labeling. In: Proceedings of the 58th Annual Meeting of the Association for Computational Linguistics. pp. 313--322 (Jul 2020). \doi{10.18653/v1/2020.acl-main.29}

\bibitem{BAYOMI18.806}
Bayomi, M., Lawless, S.: C-hts: A concept-based hierarchical text segmentation approach (may 2018)

\bibitem{chen2023chatgpts}
Chen, L., Zaharia, M., Zou, J.: How is chatgpt's behavior changing over time? (2023), \url{arXiv 2307.09009}

\bibitem{dettmers2023qlora}
Dettmers, T., Pagnoni, A., Holtzman, A., Zettlemoyer, L.: Qlora: Efficient finetuning of quantized llms (2023), \url{arXiv 2305.14314}

\bibitem{douze2024faiss}
Douze, M., Guzhva, A., Deng, C., Johnson, J., Szilvasy, G., Mazaré, P.E., Lomeli, M., Hosseini, L., Jégou, H.: The faiss library (2024), \url{arXiv 2401.08281}

\bibitem{hu2021lora}
Hu, E.J., Shen, Y., Wallis, P., Allen-Zhu, Z., Li, Y., Wang, S., Wang, L., Chen, W.: Lora: Low-rank adaptation of large language models (2021), \url{arXiv 2106.09685}

\bibitem{jin2024comprehensive}
Jin, R., Du, J., Huang, W., Liu, W., Luan, J., Wang, B., Xiong, D.: A comprehensive evaluation of quantization strategies for large language models (2024), \url{arXiv 2402.16775}

\bibitem{Karaa2011NamedER}
Karaa, W.B.A.: Named entity recognition using web document corpus. MatSciRN: Other Electronic  (2011), \url{https://api.semanticscholar.org/CorpusID:9633141}

\bibitem{kojima2023large}
Kojima, T., Gu, S.S., Reid, M., Matsuo, Y., Iwasawa, Y.: Large language models are zero-shot reasoners (2023), \url{arXiv 2205.11916}

\bibitem{poderjudicial}
Poder judicial de la nación argentina, \url{https://www.pjn.gov.ar/}

\bibitem{MiniLM}
Reimers, N., Gurevych, I.: Código fuente paraphrase-multilingual-minilm-l12-v2 huggingface., \url{https://huggingface.co/sentence-transformers/paraphrase-multilingual-MiniLM-L12-v2}

\bibitem{ROUMELIOTIS2024100056}
Roumeliotis, K.I., Tselikas, N.D., Nasiopoulos, D.K.: Llms in e-commerce: A comparative analysis of gpt and llama models in product review evaluation. Natural Language Processing Journal  \textbf{6},  100056 (2024). \doi{10.1016/j.nlp.2024.100056}

\bibitem{PyPDF2}
Thoma, M.: Pypdf2, \url{https://pypi.org/project/PyPDF2/}

\bibitem{touvron2023llama}
Touvron, H., Martin, L., Stone, K., et~al.: Llama 2: Open foundation and fine-tuned chat models (2023), \url{arXiv 2307.09288}

\bibitem{Vardhan}
Vardhan, H., Surana, N., Tripathy, B.K.: Named-entity recognition for legal documents. In: Hassanien, A.E., Bhatnagar, R., Darwish, A. (eds.) Advanced Machine Learning Technologies and Applications. pp. 469--479. Springer Singapore, Singapore (2021)

\bibitem{varshney2023stitch}
Varshney, N., Yao, W., Zhang, H., Chen, J., Yu, D.: A stitch in time saves nine: Detecting and mitigating hallucinations of llms by validating low-confidence generation (2023), \url{arXiv 2307.03987}

\bibitem{xu2023parameterefficient}
Xu, L., Xie, H., Qin, S.Z.J., Tao, X., Wang, F.L.: Parameter-efficient fine-tuning methods for pretrained language models: A critical review and assessment (2023), \url{arXiv 2312.12148}

\end{thebibliography}

\end{document}